\documentclass[11pt,a4paper]{article}
\usepackage[hyperref]{emnlp2018}
\usepackage{times}
\usepackage{latexsym}

\usepackage{url}

\usepackage{booktabs}
\usepackage{multirow}
\usepackage{graphicx}
\usepackage{amsmath}
\usepackage{wrapfig}

\allowdisplaybreaks

\aclfinalcopy % Uncomment this line for the final submission
\def\aclpaperid{2269} %  Enter the acl Paper ID here

\title{Improving Unsupervised Word-by-Word Translation\\with Language Model and Denoising Autoencoder}

\author{Yunsu Kim \hspace{10pt} Jiahui Geng \hspace{9pt} Hermann Ney \\
  Human Language Technology and Pattern Recognition Group \\
  RWTH Aachen University \\
  Aachen, Germany \\
  {\tt \{surname\}@cs.rwth-aachen.de} \\}

\date{}

\begin{document}
\maketitle
\begin{abstract}
  Unsupervised learning of cross-lingual word embedding offers elegant matching of words across languages, but has fundamental limitations in translating sentences. In this paper, we propose simple yet effective methods to improve word-by-word translation of cross-lingual embeddings, using only monolingual corpora but without any back-translation. We integrate a language model for context-aware search, and use a novel denoising autoencoder to handle reordering. Our system surpasses state-of-the-art unsupervised neural translation systems without costly iterative training. We also analyze the effect of vocabulary size and denoising type on the translation performance, which provides better understanding of learning the cross-lingual word embedding and its usage in translation.
\end{abstract}

\section{Introduction}

Building a machine translation (MT) system requires lots of bilingual data. Neural MT models \cite{nmt}, which become the current standard, are even more difficult to train without huge bilingual supervision \cite{nmt-6}. However, bilingual resources are still limited to some of the selected language pairs---mostly from or to English.

A workaround for zero-resource language pairs is translating via an intermediate (pivot) language. To do so, we need to collect parallel data and train MT models for source-to-pivot and pivot-to-target individually; it takes a double effort and the decoding is twice as slow.

Unsupervised learning is another alternative, where we can train an MT system with only monolingual corpora. Decipherment methods \cite{deciphering,beam-cipher} are the first work in this direction, but they often suffer from a huge latent hypothesis space \cite{largevocab}.

Recent work by \newcite{unmt-artetxe} and \newcite{unmt-facebook} train sequence-to-sequence MT models of both translation directions together in an unsupervised way. They do back-translation \cite{bt} back and forth for every iteration or batch, which needs an immensely long time and careful tuning of hyperparameters for massive monolingual data.

Here we suggest rather simple methods to build an unsupervised MT system quickly, based on word translation using cross-lingual word embeddings. The contributions of this paper are:

\begin{itemize}\itemsep0em
\item We formulate a straightforward way to combine a language model with cross-lingual word similarities, effectively considering context in lexical choices.
\item We develop a postprocessing method for word-by-word translation outputs using a denoising autoencoder, handling local reordering and multi-aligned words.
\item We analyze the effect of different artificial noises for the denoising model and propose a novel noise type.
\item We verify that cross-lingual embedding on subword units performs poorly in translation.
\item We empirically show that cross-lingual mapping can be learned using a small vocabulary without losing the translation performance.
\end{itemize}

The proposed models can be efficiently trained with off-the-shelf softwares with little or no changes in the implementation, using only monolingual data. The provided analyses help for better learning of cross-lingual word embeddings for translation purpose. Altogether, our unsupervised MT system outperforms the sequence-to-sequence neural models even without training signals from the opposite translation direction, i.e. via back-translation.

\section{Cross-lingual Word Embedding}
\label{sec:cross}
As a basic step for unsupervised MT, we learn a word translation model from monolingual corpora of each language. In this work, we exploit cross-lingual word embedding for word-by-word translation, which is state-of-the-art in terms of type translation quality \cite{cross-artetxe,cross-facebook}.

Cross-lingual word embedding is a continuous representation of words whose vector space is shared across multiple languages. This enables distance calculation between word embeddings across languages, which is actually finding translation candidates.

We train cross-lingual word embedding in a fully unsupervised manner:

\begin{enumerate}\itemsep0em
\item \label{step:mono-embed} Learn monolingual source and target embeddings independently. For this, we run skip-gram algorithm augmented with character $n$-gram \cite{fasttext}.
\item Find a linear mapping from source embedding space to target embedding space by adversarial training \cite{cross-facebook}. We do not pre-train the discriminator with a seed dictionary, and consider only the top $V_\text{cross-train}$ words of each language as input to the discriminator.
\end{enumerate}

Once we have the cross-lingual mapping, we can transform the embedding of a given source word and find a target word with the closest embedding, i.e. nearest neighbor search. Here, we apply cross-domain similarity local scaling \cite{cross-facebook} to penalize the word similarities in dense areas of the embedding distribution.

We further refine the mapping obtained from Step 2 as follows \cite{cross-artetxe}:
\begin{enumerate}\itemsep0em
  \setcounter{enumi}{2}
\item Build a synthetic dictionary by finding mutual nearest neighbors for both translation directions in vocabularies of $V_\text{cross-train}$ words.
\item Run a Procrustes problem solver with the dictionary from Step 3 to re-train the mapping \cite{smith}.
\item Repeat Step 3 and 4 for a fixed number of iterations to update the mapping further.
\end{enumerate}

\section{Sentence Translation}

In translating sentences, cross-lingual word embedding has several drawbacks. We describe each of them and our corresponding solutions.

\subsection{Context-aware Beam Search}
\label{sec:context}
The word translation using nearest neighbor search does not consider context around the current word. In many cases, the correct translation is not the nearest target word but other close words with morphological variations or synonyms, depending on the context.% For example, a German word ``\textit{verstehen}'', which means ``to understand'' in most cases, has the following top-5 nearest neighbors in English vocabulary in a cross-lingual embedding space:

% \begin{table}[!ht]
%   \centering
%   \begin{tabular}{cc}
%     \toprule
%     Word & Cosine Similarity \\
%     \midrule
%     \textit{understand} & 0.664 \\
%     \textit{verbalize} & 0.511 \\
%     \textit{explain} & 0.508 \\
%     \textit{understanding} & 0.507 \\
%     \textit{comprehend} & 0.506 \\
%     \bottomrule
%   \end{tabular}
%   \caption{Nearest neighbors of German word \textit{verstehen}.}
%   \label{tab:nn}
% \end{table}

% The list has also a nominalized form (``\textit{understanding}'') and a synonym (``\textit{comprehend}'') of the top candidate ``\textit{understand}''. A correct translation of ``\textit{verstehen}'' can be among these, depending on the context. We also observe clearly wrong translations (``\textit{verbalize}'', ``\textit{explain}''), where context helps to exclude them by giving a lower score.

The reasons are in two-fold: 1) Word embedding is trained to place semantically related words nearby, even though they have opposite meanings. 2) A hubness problem of high-dimensional embedding space hinders a correct search, where lots of different words happen to be close to each other \cite{hubness}.

In this paper, we integrate context information into word-by-word translation by combining a language model (LM) with cross-lingual word embedding. Let $f$ be a source word in the current position and $e$ a possible target word. Given a history $h$ of target words before $e$, the score of $e$ to be the translation of $f$ would be: 
\vspace{-0.2em}
\begin{align*}
  L(e;f,h) = \lambda_\text{emb}\log q(f,e) + \lambda_\text{LM}\log p(e|h)
\end{align*}

Here, $q(f,e)$ is a lexical score defined as:
\vspace{-0.2em}
\begin{align*}
  q(f,e) = \frac{d(f,e) + 1}{2}
\end{align*}
where $d(f,e) \in [-1,1]$ is a cosine similarity between $f$ and $e$. It is transformed to the range $[0,1]$ to make it similar in scale with the LM probability. In our experiments, we found that this simple linear scaling is better than sigmoid or softmax functions in the final translation performance.

Accumulating the scores per position, we perform a beam search to allow only reasonable translation hypotheses.

\subsection{Denoising}

Even when we have correctly translated words for each position, the output is still far from an acceptable translation. We adopt sequence denoising autoencoder \cite{sdae} to improve the translation output of Section \ref{sec:context}. The main idea is to train a sequence-to-sequence neural network model that takes a noisy sentence as input and produces a (denoised) clean sentence as output, both of which are of the same (target) language. The model was originally proposed to learn sentence embeddings, but here we use it directly to actually remove noise in a sentence.

Training label sequences for the denoising network would be target monolingual sentences, but we do not have their noisy versions at hand. Given a clean target sentence, the noisy input should be ideally word-by-word translation of the corresponding source sentence. However, such bilingual sentence alignment is not available in our unsupervised setup.

Instead, we inject artificial noise into a clean sentence to simulate the noise of word-by-word translation. We design different noise types after the following aspects of word-by-word translation.

\subsubsection{Insertion}

\begin{figure}[!ht]
  \centering
  \hspace{0.5cm}\includegraphics[width=0.8\linewidth]{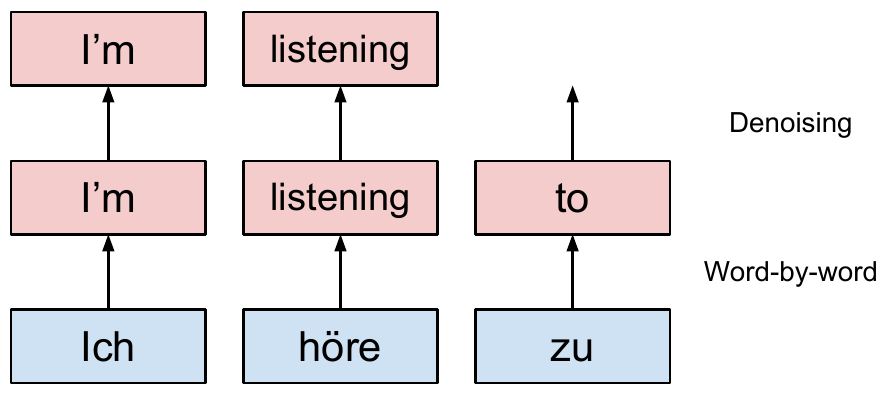}
  \caption{Example of denoising an insertion noise.}
  \label{fig:insertion}
\end{figure}

Word-by-word translation always outputs a target word for every position. However, there are a plenty of cases that multiple source words should be translated to a single target word, or that some source words are rather not translated to any word to make a fluent output. For example, a German sentence ``\textit{Ich h\"{o}re zu.}'' would be translated to ``\textit{I'm listening to.}'' by a word-by-word translator, but ``\textit{I'm listening.}'' is more natural in English (Figure \ref{fig:insertion}).

%One might think of having an empty token in the target vocabulary to handle this problem, generating this token for suitable positions. To implement this, however, we need a target corpus including empty tokens for learning to predict such positions. Whether to put an empty token or not can only be learned from word alignments of bilingual sentence pairs, which is again not available in unsupervised MT.

We pretend to have extra target words which might be translation of redundant source words, by inserting random target words to a clean sentence:

\begin{enumerate}\itemsep0em
\item For each position $i$, sample a probability $p_i \sim \text{Uniform}(0,1)$.
\item If $p_i < p_\text{ins}$, sample a word $e$ from the most frequent $V_\text{ins}$ target words and insert it before position $i$.
\end{enumerate}

We limit the inserted words by $V_\text{ins}$ because target insertion occurs mostly with common words, e.g. prepositions or articles, as the example above. We insert words only before---not after---a position, since an extra word after the ending word (usually a punctuation) is not probable.

\subsubsection{Deletion}

\begin{figure}[!ht]
  \centering
  \includegraphics[width=\linewidth]{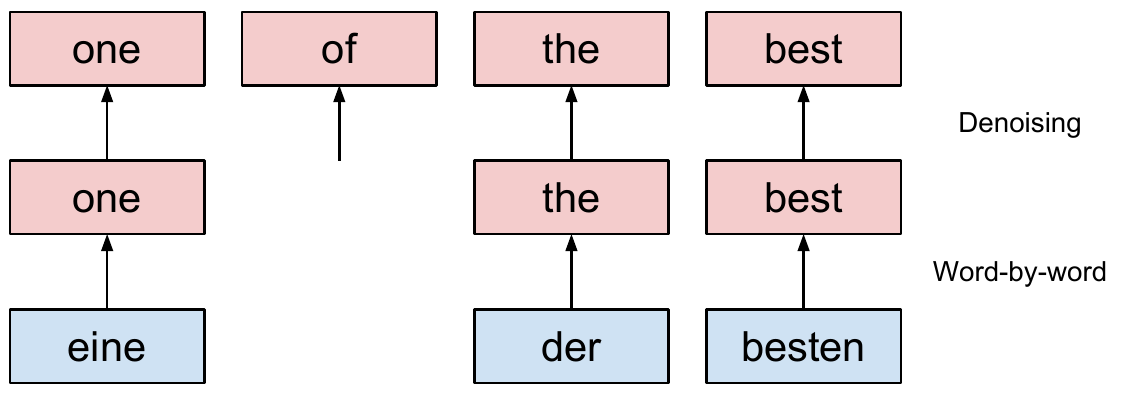}
  \caption{Example of denoising a deletion noise.}
  \label{fig:deletion}
\end{figure}

Similarly, word-by-word translation cannot handle the contrary case: when a source word should be translated into more than one target words, or a target word should be generated from no source words for fluency. For example, a German word ``\textit{im}'' must be ``\textit{in the}'' in English, but word translation generates only one of the two English words. Another example is shown in Figure \ref{fig:deletion}.

To simulate such situations, we drop some words randomly from a clean target sentence \cite{sdae}:

\begin{enumerate}\itemsep0em
\item For each position $i$, sample a probability $p_i \sim \text{Uniform}(0,1)$.
\item If $p_i < p_\text{del}$, drop the word in the position $i$.
\end{enumerate}

\subsubsection{Reordering}
\begin{figure}[!ht]
  \centering
  \includegraphics[width=\linewidth]{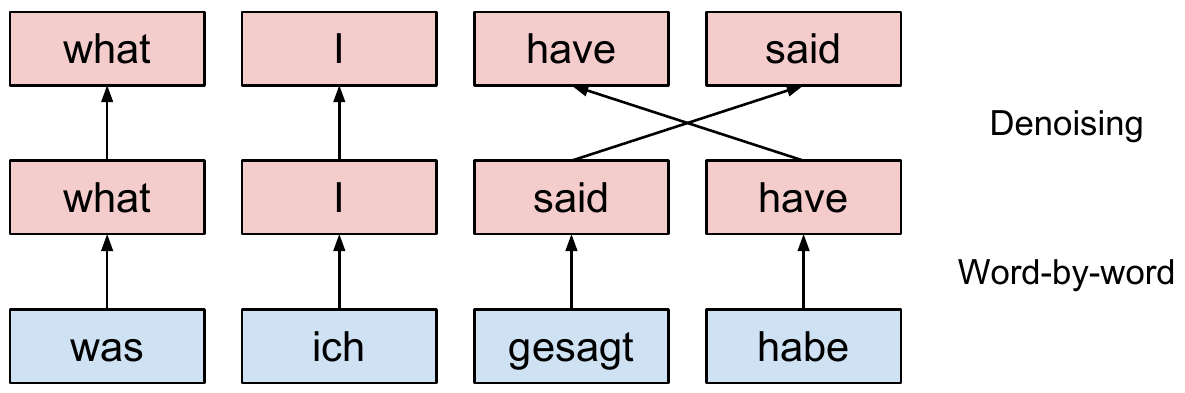}
  \caption{Example of denoising the reordering noise.}
  \label{fig:reordering}
\end{figure}

Also, translations generated word-by-word are not in an order of the target language. In our beam search, LM only assists in choosing the right word in context but does not modify the word order. A common reordering problem of German$\rightarrow$English is illustrated in Figure \ref{fig:reordering}.

From a clean target sentence, we corrupt its word order by random permutations. We limit the maximum distance between an original position and its new position like \newcite{unmt-facebook}:

\begin{enumerate}\itemsep0em
\item For each position $i$, sample an integer $\delta_i$ from $[0,d_\text{per}]$.
\item Add $\delta_i$ to index $i$ and sort the incremented indices $i + \delta_i$ in an increasing order.
\item Rearrange the words to be in the new positions, to which their original indices have moved by Step 2.
\end{enumerate}

% Main result table for section 4, moved here for layout
\begin{table*}[!ht]
  \centering
  \begin{tabular}{lcccc}
    \toprule
    & de-en & en-de & fr-en & en-fr\\
    System & \textsc{Bleu} [\%] & \textsc{Bleu} [\%] & \textsc{Bleu} [\%] & \textsc{Bleu} [\%]\\
    \midrule
    Word-by-Word & 11.1 & 6.7 & 10.6 & 7.8\\
    + LM & 14.5 & 9.9 & 13.6 & 10.9\\
    \hspace{10pt}+ Denoising & \textbf{17.2} & \textbf{11.0} & \textbf{16.5} & 13.9 \\
    \midrule
    \cite{unmt-facebook} & 13.3 & 9.6 & 14.3 & 15.1\\
    \cite{unmt-artetxe} & - & - & 15.6 & 15.1\\
    \bottomrule
  \end{tabular}
  \caption{Translation results on German$\leftrightarrow$English \texttt{newstest2016} and French$\leftrightarrow$English \texttt{newstest2014}.}
  \label{tab:results}
\end{table*}

This is a generalized version of swapping two neighboring words \cite{sdae}. Reordering is highly dependent of each language, but we found that this noise is generally close to word-by-word translation outputs.\\

\noindent Insertion, deletion, and reordering noises were applied to each mini-batch with different random seeds, allowing the model to see various noisy versions of the same clean sentence over the epochs.

Note that the deletion and permutation noises are integrated in the neural MT training of \newcite{unmt-artetxe} and \newcite{unmt-facebook} as additional training objectives. Whereas we optimize an independent model solely for denoising without architecture change. It allows us to easily train a larger network with a larger data. Insertion noise is of our original design, which we found to be the most effective (Section \ref{sec:study-noise}).

\section{Experiments}
We applied the proposed methods on WMT 2016 German$\leftrightarrow$English task and WMT 2014 French$\leftrightarrow$English task. For German/English, we trained word embeddings with 100M sentences sampled from News Crawl 2014-2017 monolingual corpora. For French, we used News Crawl 2007-2014 (around 42M sentences). The data was lowercased and filtered to have a maximum sentence length 100. German compound words were splitted beforehand. Numbers were replaced with category labels and recovered back after decoding by looking at the source sentence.

fasttext \cite{fasttext} was used to learn monolingual embeddings for only the words with minimum count 10. MUSE \cite{cross-facebook} was used for cross-lingual mappings with $V_\text{cross-train}$ = 100k and 10 refinement iterations (Step 3-5 in Section \ref{sec:cross}). Other parameters follow the values in \newcite{cross-facebook}. With the same data, we trained 5-gram count-based LMs using KenLM \cite{kenlm} with its default setting.

Denoising autoencoders were trained using Sockeye \cite{sockeye} on News Crawl 2016 for German/English and News Crawl 2014 for French. We considered only top 50k frequent words for each language and mapped other words to \texttt{<unk>}. The unknowns in the denoised output were replaced with missing words from the noisy input by a simple line search.

We used 6-layer Transformer encoder/decoder \cite{transformer} for denoisers, with embedding/hidden layer size 512, feedforward sublayer size 2048 and 8 attention heads.

As a validation set for the denoiser training, we used \texttt{newstest2015} (German $\leftrightarrow$ English) or \texttt{newstest2013} (French $\leftrightarrow$ English), where the input/output sides both have the same clean target sentences, encouraging a denoiser to keep at least clean part of word-by-word translations. Here, the noisy input showed a slight degradation of performance; the model seemed to overfit to specific noises in the small validation set.

Optimization of the denoising models was done with Adam \cite{adam}: initial learning rate 0.0001, checkpoint frequency 4000, no learning rate warmup, multiplying 0.7 to the learning rate when the perplexity on the validation set did not improve for 3 checkpoints. We stopped the training if it was not improved for 8 checkpoints.

In decoding, we used $\lambda_\text{embed} = 1$ and $\lambda_\text{LM} = 0.1$ with beam size 10. We only translated top frequent 50k source words and merely copied other words to target side. For each position, only the nearest 100 target words were considered.

Table \ref{tab:results} shows the results. LM improves word-by-word baselines consistently in all four tasks, giving at least +3\% \textsc{Bleu}. When our denoising model is applied on top of it, we have additional gain around +3\% \textsc{Bleu}. Note that our methods do not involve any decoding steps to generate pseudo-parallel training data, but still perform better than unsupervised MT systems that rely on repetitive back-translations \cite{unmt-artetxe,unmt-facebook} by up to +3.9\% \textsc{Bleu}. The total training time of our method is only 1-2 days with a single GPU.
% Here we also compare different strategies for handling source unknowns. We verify that copying unknown words is more effective: +1-1.5\% \textsc{Bleu} better than choosing translations among words with high LM probabilities.

\subsection{Ablation Study: Denoising}
\label{sec:study-noise}
\begin{table}[!ht]
  \centering
  \begin{tabular}{ccrc}
    \toprule
    $d_\text{per}$ & $p_\text{del}$ & $V_\text{ins}$ & \textsc{Bleu} [\%] \\
    \midrule
    2 & & & 14.7\\
    3 & & & \textbf{14.9}\\
    5 & & & 14.9\\
    \midrule
    \multirow{2}{*}{3} & 0.1 & & \textbf{15.7} \\
      & 0.3 & & 15.1 \\
    \midrule
    \multirow{4}{*}{3} & \multirow{4}{*}{0.1} & 10 & 16.8 \\
                   & & 50 & \textbf{17.2} \\
                   & & 500 & 16.8 \\
                   & & 5000 & 16.5\\
    \bottomrule
  \end{tabular}
  \caption{Translation results with different values of denoising parameters for German$\rightarrow$English.}
  \label{tab:denoising}
\end{table}

To examine the effect of each noise type in denoising autoencoder, we tuned each parameter of the noise and combined them incrementally (Table \ref{tab:denoising}). Firstly, for permutations, a significant improvement is achieved from $d_\text{per} = 3$, since a local reordering usually involves a sequence of 3 to 4 words. With $d_\text{per} > 5$, it shuffles too many consecutive words together, yielding no further improvement. This noise cannot handle long-range reordering, which is usually a swap of words that are far from each other, keeping the words in the middle as they are.

Secondly, we applied the deletion noise with different values of $p_\text{del}$. 0.1 gives +0.8\% \textsc{Bleu}, but we immediately see a degradation with a larger value; it is hard to observe one-to-many translations more than once in each sentence pair.

Finally, we optimized $V_\text{ins}$ for the insertion noise, fixing $p_\text{ins} = 0.1$. Increasing $V_\text{ins}$ is generally not beneficial, since it provides too much variations in the inserted word; it might not be related to its neighboring words. Overall, we observe the best result (+1.5\% \textsc{Bleu}) with $V_\text{ins} = 50$.

\subsection{Ablation Study: Vocabulary}
\label{sec:vocab}
\begin{table}[!ht]
  \centering
  \begin{tabular}{lcc}
    \toprule
    \multicolumn{2}{c}{Vocabulary} & \textsc{Bleu} [\%] \\
    \midrule
                                   & Merges \\
    \cmidrule{2-2}
    \multirow{3}{*}{BPE} & 20k & 10.4 \\
    & 50k & 12.5 \\
    & 100k & 13.0 \\
    \midrule
    & $V_\text{cross-train}$\\
    \cmidrule{2-2}
    \multirow{4}{*}{Word} & 20k & 14.4\\
    & 50k & 14.4\\
    & 100k & \textbf{14.5}\\
    & 200k & 14.4\\
    \bottomrule
  \end{tabular}
  \caption{Translation results with different vocabularies for German$\rightarrow$English (without denoising).}
  \label{tab:vocab}
  \vspace{-0.5em}
\end{table}

We also examined how the translation performance varies with different vocabularies of cross-lingual word embedding in Table \ref{tab:vocab}. The first three rows show that BPE embeddings perform worse than word embeddings, especially with smaller vocabulary size. For short BPE tokens, the context they meet during the embedding training is much more various than a complete word, and a direct translation of such token to a BPE token of another language would be very ambiguous.

For word level embeddings, we compared different vocabulary sizes used for training the cross-lingual mapping (the second step in Section \ref{sec:cross}). Surprisingly, cross-lingual word embedding learned only on top 20k words is comparable to that of 200k words in the translation quality. We also increased the search vocabulary to more than 200k but the performance only degrades. This means that word-by-word translation with cross-lingual embedding depends highly on the frequent word mappings, and learning the mapping between rare words does not have a positive effect.

\section{Conclusion}

In this paper, we proposed a simple pipeline to greatly improve sentence translation based on cross-lingual word embedding. We achieved context-aware lexical choices using beam search with LM, and solved insertion/deletion/reordering problems using denoising autoencoder. Our novel insertion noise shows a promising performance even combined with other noise types. Our methods do not need back-translation steps but still outperforms costly unsupervised neural MT systems. In addition, we proved that for general translation purpose, an effective cross-lingual mapping can be learned using only a small set of frequent words, not on subword units. Our implementation of the LM integration\footnote{\scriptsize \url{https://github.com/yunsukim86/wbw-lm}} and the denoising autoencoder\footnote{\scriptsize \url{https://github.com/yunsukim86/sockeye-noise}} is available online.

%As future work, we plan to apply the same methods on monolingually learned phrases. With carefully selected phrases, local reordering problem will be directly solved by phrase-by-phrase translation. Also, our denoising model will be able to deal with phrase level swaps.

\section*{Acknowledgments}
\begin{wrapfigure}{l}{0.18\textwidth}
  \centering
  \vspace{-0.5em}
  \includegraphics[width=0.2\textwidth]{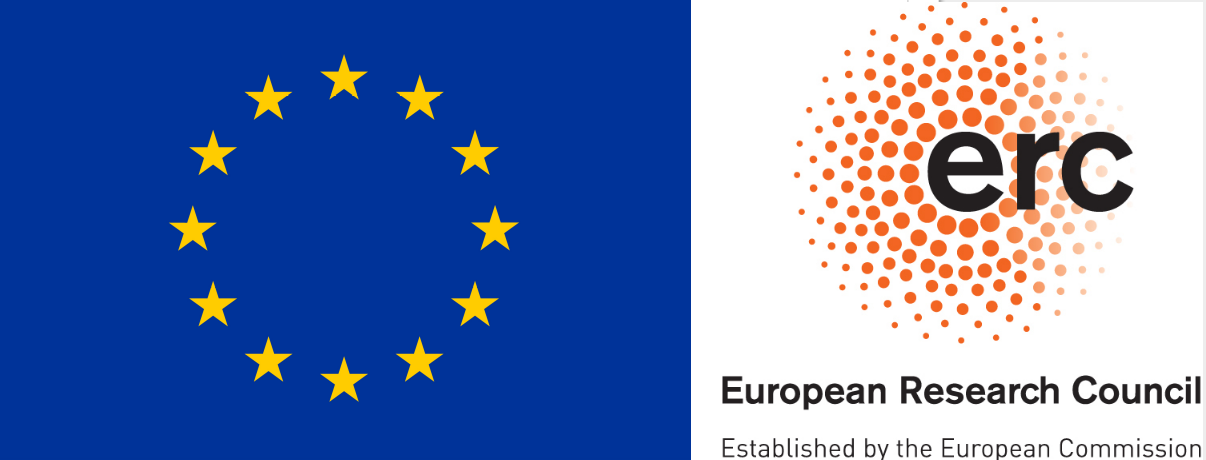}\\[0.6em]
  \hspace{11pt}\includegraphics[width=0.15\textwidth]{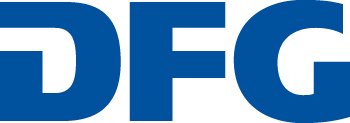}
  \vspace{-1.5em}
\end{wrapfigure}
This work has received funding from the European Research Council (ERC) under the European Union's Horizon 2020 research and innovation programme, grant agreement No. 694537 (SEQCLAS). The GPU computing cluster was partially funded by Deutsche Forschungsgemeinschaft (DFG) under grant INST 222/1168-1 FUGG. The work reflects only the authors' views and neither ERC nor DFG is responsible for any use that may be made of the information it contains.
\vspace{-0.3em}
\bibliography{references}
\bibliographystyle{acl_natbib_nourl}

\end{document}